\documentclass{article}

\usepackage{arxiv}

\usepackage[utf8]{inputenc} 
\usepackage[T1]{fontenc}    
\usepackage{hyperref}       
\usepackage{url}            
\usepackage{booktabs}       
\usepackage{amsfonts}       
\usepackage{nicefrac}       
\usepackage{microtype}      
\usepackage{natbib}
\usepackage{cleveref}
\usepackage{graphicx}
\usepackage[
  separate-uncertainty = true,
  multi-part-units = repeat
]{siunitx}
\usepackage[section]{placeins}
\usepackage{subcaption}

\title{Path Capsule Networks}

\author{
  Mohammed Amer\thanks{Corresponding author} \\
  School of Computer Science\\
  University of Nottingham\\
  Semenyih, Malaysia \\
  \texttt{hcxma1@nottingham.edu.my} \\
   \And
  Tom\'as Maul\\
  School of Computer Science\\
  University of Nottingham\\
  Semenyih, Malaysia \\
  \texttt{tomas.maul@nottingham.edu.my} \\
}

\begin{document}
\maketitle

\begin{abstract}

Capsule network (CapsNet) was introduced as an enhancement over convolutional neural networks, supplementing the latter's invariance properties with equivariance through pose estimation. CapsNet achieved a very decent performance with a shallow architecture and a significant reduction in parameters count. However, the width of the first layer in CapsNet is still contributing to a significant number of its parameters and the shallowness may be limiting the representational power of the capsules. To address these limitations, we introduce Path Capsule Network (PathCapsNet), a deep parallel multi-path version of CapsNet. We show that a judicious coordination of depth, max-pooling, regularization by DropCircuit and a new fan-in routing by agreement technique can achieve better or comparable results to CapsNet, while further reducing the parameter count significantly.

\end{abstract}

\section{Introduction}

Convolutional Neural Networks (CNNs) \citep{fukushima1980,lecun1995} have remained state-of-the-art in image processing and computer vision tasks since their successful large scale training by \citet{Krizhevsky2012}. CNNs were biologically inspired by the visual cortex \citep{Hubel1968} and were built on the principle of translation invariance, achieved through local receptive fields, weight sharing and pooling operations. Despite their success, CNNs suffer from inherent limitations, most significantly the fact that translation invariance by definition causes loss of location information. This limitation has stimulated a lot of research in the direction of augmenting learning with location data \citep{Wang2018, Tang2015, Ghafoorian2017}. 

\citet{Sabour2017} argued that the main limitation of CNNs is the focus on achieving translation invariance, and that equivariance should also be targeted. Hence, the authors proposed CapsNet as a step towards achieving equivariance. The philosophy of CapsNet is that a single activation/feature should be replaced by a pose vector, named capsule, representing the different properties of an object's viewpoint. CapsNet has two main components, which are PrimaryCapsule and DigitCaps layers. PrimaryCapsules represent the different parts of the underlying objects, which are then multiplied by translation matrices to get prediction vectors, representing the votes of each PrimaryCapsule with respect to each DigitCaps, which are then routed using routing by agreement to compute DigitCaps activations, which can then be used to signify the presence of an object. The philosophy is that with changing the viewpoint of an object, the change in pose matrices should be coordinated, such that the voting agreement is maintained. We consider using another form of routing by agreement, fan-in routing in contrast to fan-out routing used by \citet{Sabour2017}, which we show can have better performance under some conditions.

CapsNet was shown to achieve very good results with a shallow architecture and decent parameter savings, compared to deep CNNs. However, the lack of depth can be limiting to the expressiveness of the network. Moreover, the first convolutional layer in CapsNet is large and contributes to increasing the number of CapsNet parameters significantly. We believe that a coordinated inclusion of depth and multiple pathways can help increase the network performance and simultaneously help save more parameters.

We consider a multipath architecture for including more depth into CapsNet. Multiple paths in neural networks are biologically plausible and biological neural networks have been shown to exhibit multipath parallel processing \citep{Gollisch2010, Otsuna2014}. Aside from biological inspiration, we think that using different paths for generating PrimaryCapsules can be exploited to enhance performance while saving parameters significantly. A PrimaryCapsule generated by a deep path can be considered a deep version of the original CapsNet capsule, which we believe can exhibit more expressiveness and more abstraction, similar to other deep structures in the deep learning paradigm.

The universal approximation theorem by \citet{Hornik1990} showed that any Borel measurable function can be approximated by a sufficiently wide single layer multilayer perceptron (MLP). Empirically, however, this is infeasible due to optimization limitations, and is rarely desirable due to the problem of overfitting. On the other hand, making use of depth is statistically motivated by composition of functions and empirically can lead to better generalization. Moreover, as we show, depth can be added judiciously to save parameters without sacrificing performance.

Our contributions in this paper are:

\begin{enumerate}
    \item We propose PathCapsNet, a multipath deep version of CapsNet.
    \item We enrich the routing by agreement methodology by a new variant, fan-in routing.
    \item By carefully adding depth and max-pooling, along with a multi-path structure, fan-in routing and DropCircuit, we achieved comparable results to CapsNet with significant parameter savings.
    \item We open the possibility of leveraging significant model parallelism in the context a capsule networks.
\end{enumerate}

In the next section we discuss the previous work done around capsule networks and multipath architectures, and how we enhance by building on these concepts.

\section{Related Work}

CapsNet \citep{Sabour2017} was introduced as an architecture that builds up on the conventional CNN \citep{fukushima1980,LeCun1989} trying to overcome its limitations. The main motivation behind CapsNet is achieving equivariance, in addition to the invariance properties already implemented by CNN. CapsNet could achieve a good generalization using relatively fewer parameters than deep CNNs (only 8.2M parameters for the MNIST model with reconstruction). Different variants have been introduced since the original CapsNet. \citet{Phaye2018} introduced DCNet as a dense version of capsule networks and DCNet++ by stacking multiple DCNets. In DCNet++, each DCNet in the stack produces its version of the PrimaryCapsule layer, which is then fed to the next DCNet in the stack. The final output is calculated based on both the output of each subnetwork and their concatenation. They also made some modifications to the decoder (reconstruction) subnetwork. DCNet++ achieved good generalization in relatively few epochs at the cost of using more parameters (13.4M).

Another variant is MS-CapsNet \citep{Xiang2018}. MS-CapsNet is composed of three successive modules. The first module is the feature extractor and it has two convolutional paths of depths 1 and 2 and a third path which is just a skip connection. Each path produces a PrimaryCapsule of different dimension. The second module is a capsule encoding and it is responsible for projecting the PrimaryCapsules to a common dimension and concatenating them. The third module, capsule dropout, is applied before routing and it is responsible for dropping random capsules as a way of regularization in a manner similar to dropout \citep{Srivastava2014} and other similar techniques. Capsule dropout showed enhancement in performance relative to the non-dropout condition. MS-CapsNet could achieve better performance than the original CapsNet on FashionMNIST and CIFAR-10 with fewer parameters ($\sim$11M).

SECaps \citep{He2018} is an adaptation of CapsNet to sequential tasks, specifically Natural Language Processing (NLP). The word embeddings of single words are treated as PrimaryCapsules. Since the dynamic routing is not sequential in nature and doesn't respect order, the seq-caps layer is introduced. This layer is basically composed of a long short-term memory (LSTM) layer that is applied to a given sequence of the data as a series encoding, and then the output is dynamically routed in the conventional way to produce the output of the next layer. Multiple seq-caps layers can be stacked. Another module, the attention module, transforms the word embeddings, which are then concatenated with the seq-layer output. The final output is produced by an MLP subnetwork. SECaps was evaluated on multiple charge prediction datasets, achieving better performance than the state-of-the-art.

Siamese capsule network (SCN) \citep{Neill2018} is the capsule version of the conventional siamese network. \citet{Neill2018} introduced SCN as a face verification approach similar to DeepFace \citep{Taigman2014}. SCN is very similar in architecture to the original CapsNet. It has a convolutional layer, followed by the PrimaryCapsules layer and then a layer called Face Capsule layer, which is essentially similar to the DigitCaps layer. The final output is produced by a fully connected layer on top of the Face Capsule layer. SCN achieved good performance on different datasets with a smaller model, little preprocessing and less data. 

Matrix capsules network was proposed by \citet{Hinton2018} as a generalization of the original CapsNet for more efficient pose estimation. Each capsule is represented by a matrix and a sigmoid unit that controls the probability of activating the capsule. Every pose matrix is multiplied by a transformation matrix to get the votes which will be used for routing to the next layer. Routing is done using expectation maximization (EM) that takes as input the votes and activation probabilities of the previous layer. Matrix capsules network achieved a very good accuracy improvement on the smallNORB dataset, a dataset that is highly viewpoint variant, but it seems that it doesn't have the same advantage on MNIST.

The ideas of branching, parallel computation and multiple paths are well established in the deep learning literature and have their supporting biological plausibility  \citep{Gollisch2010,Otsuna2014}. In \citep{Ciresan2012}, each path in a multi-path CNN is trained on a different preprocessing/distortion of the input image and the columns outputs are averaged to produce the final output. A similar approach is used in \citep{Wang2015}, but with different types of inputs which are the source image and a bilateral filtered version of it, and the outputs of the paths are integrated using fully connected layers. \citet{Szegedy2015a} proposed the Inception-v1 model , which was responsible for winning ILSVRC-14, and is composed of a highly branched multipath architecture. \citet{Szegedy2015b} further improved the design of Inception-v1 to produce Inception-v2\&3 which exploit large scale branching and multiple paths even more. The Xception architecture \citep{Chollet2016} is a further extension to the Inception family, that uses more branching based on separable convolutions. ResNetXt \citep{Xie2016} and Residual Inception \citep{Zhang2018} are extensions of ResNet \citep{He2016} where the modular block is multipath instead of single path. 

FractalNet \citep{Larsson2016} is another type of architecture that has a recursive self-similar, highly branched structure. Parallel circuit networks, introduced by \citet{Phan2016}, adopt an extensively multipath architecture, and have demonstrated generalization  improvements using a dropping technique called DropCircuit \citep{Phan2018}. Related to the DropCircuit technique is the path dropout used by \citet{Bender2018} to regularize the training of a one-shot model, which is an implicit form of a multipath network, where a whole space of possible branches is trained simultaneously.  

We build on previous work by:

\begin{enumerate}
    \item Adding representational power to PrimaryCapsules by generating each capsule using a deep path.
    \item Enriching dynamic routing by agreement with a new fan-in variant.
    \item Combining depth, a multipath architecture, DropCircuit, max-pooling and fan-in routing to obtain a level of performance congruent with the original CapsNet, with significant parameter savings.
    \item Showing that max-pooling is not inherently contradictory with the CapsNet philosophy, and that it can be used to save parameters significantly without sacrificing nether performance nor pose awareness.
\end{enumerate}

In the next section, we explain the general PathCapsNet architecture and the different pieces that contribute to its performance.

\section{Methods}

The original CapsNet \citep{Sabour2017} has two main capsule types, namely the PrimaryCapsules and the DigitCaps. PrimaryCapsules are formed by applying an initial convolution layer to produce 256 channels, then another set of convolutions, which are then rearranged into 32 8D PrimaryCapsules. PrimaryCapsules are then routed to the next DigitCaps layer using dynamic routing by agreement. In one variant of CapsNet, namely CapsNet with reconstruction, a reconstruction layer is learned on top of the DigitCaps layer to facilitate the learning of instantiation (or transformation) parameters and therefore enhance generalization.

\begin{figure}[h]
    \centering
    \includegraphics[width=0.5\textwidth]{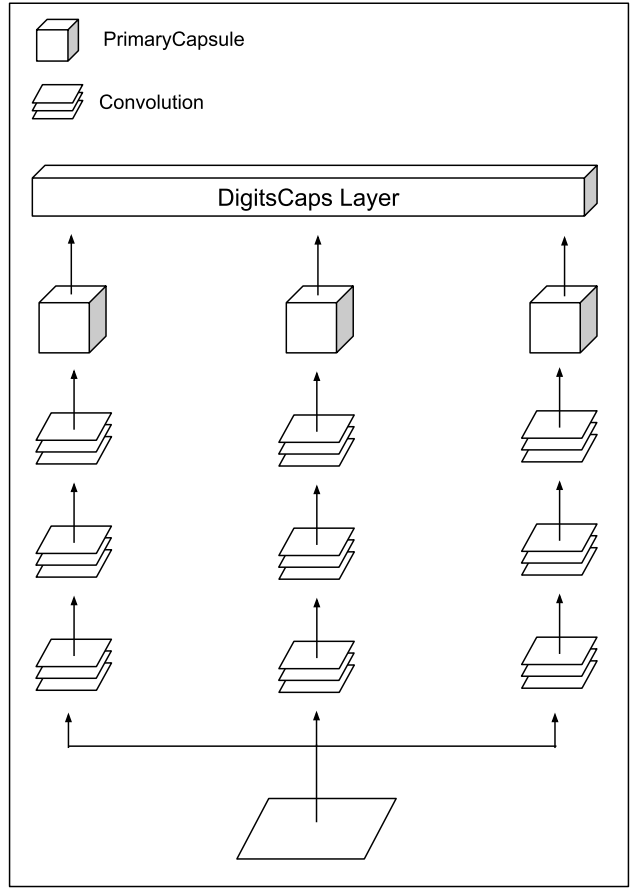}
    \caption{PathCapsNet architecture}
    \label{fig:pcnet}
\end{figure}

PathCapsNet \cref{fig:pcnet} shares the upper part of CapsNet, starting from the PrimaryCapsules layer, through the DigitCaps layer and ending with a reconstruction layer if needed. However, PathCapsNet is fundamentally different in how the PrimaryCapsules are constructed. In PathCapsNet, each PrimaryCapsule is formed by a deep CNN, named a path. So, the input is fed into different CNNs (paths) and the output of each path comprises one PrimaryCapsule.

The experiments done by \citet{Phan2018} demonstrate enhanced generalization in multipath MLPs, named parallel circuits in their work, using a drop technique called DropCircuit. DropCircuit is an adaptation of dropout to multipath architectures, where different paths are dropped during training, using a pre-specified probability. This is believed to enhance generalization by promoting independence between paths, hence allowing for problem decomposition and learning more useful representations, similar to dropout \citep{Srivastava2014} and related techniques. 

Dynamic routing is the mechanism by which PrimaryCapsules are routed to DigitCaps capsules, such that similar votes from PrimaryCapsules contribute more strongly to the target DigitCaps. The dynamic routing by agreement algorithm used in \citep{Sabour2017} updates the contribution of votes based on the similarity between the output DigitCaps and the prediction vector, representing the vote, using dot product as a measure of similarity. So, given the prediction vectors (votes) from the previous layer of capsules (PrimaryCapsule layer) $\hat{\bf{u_{j|i}}}$, where $j$ is the index of the DigitCaps capsule and $i$ is the index of a single capsule in the PrimaryCapsule layer, the output vector (DigitCaps) is calculated as,

\begin{equation}
    \textbf{s}_j = \sum_i{c_{ij}\hat{\textbf{u}}_{j|i}}
\end{equation}

where $c_{ij}$ are the coupling coefficients weighting the contributions of different prediction vectors,

\begin{equation}
    c_{ij}^{(fout)} = \frac{\exp(b_{ij})}{\sum_k{\exp(b_{ik})}}
\end{equation}

and $b_{ij}$ is the log probability (logits) that the $i$th PrimaryCapsule should be coupled to the $j$th DigitCaps capsule. We call this fan-out (fout) routing, since the weights of the contributions of the $i$th PrimaryCapsule to each DigitCaps capsule in the next layer are normalized probabilities that sum to $1.0$. For PathCapsNet, we used a different form of dynamic routing by agreement, named fan-in (fin) routing, where logits are normalized such that the weights of the contributions to the $j$th DigitCaps capsule from all the PrimaryCapsules are normalized probabilities that sum to $1.0$. Coupling coefficients for fan-in routing are calculated as,

\begin{equation}
    c_{ij}^{(fin)} = \frac{\exp(b_{ij})}{\sum_k{\exp(b_{ki})}}
\end{equation}

\begin{figure}[h]
    \centering
    \includegraphics[width=\textwidth]{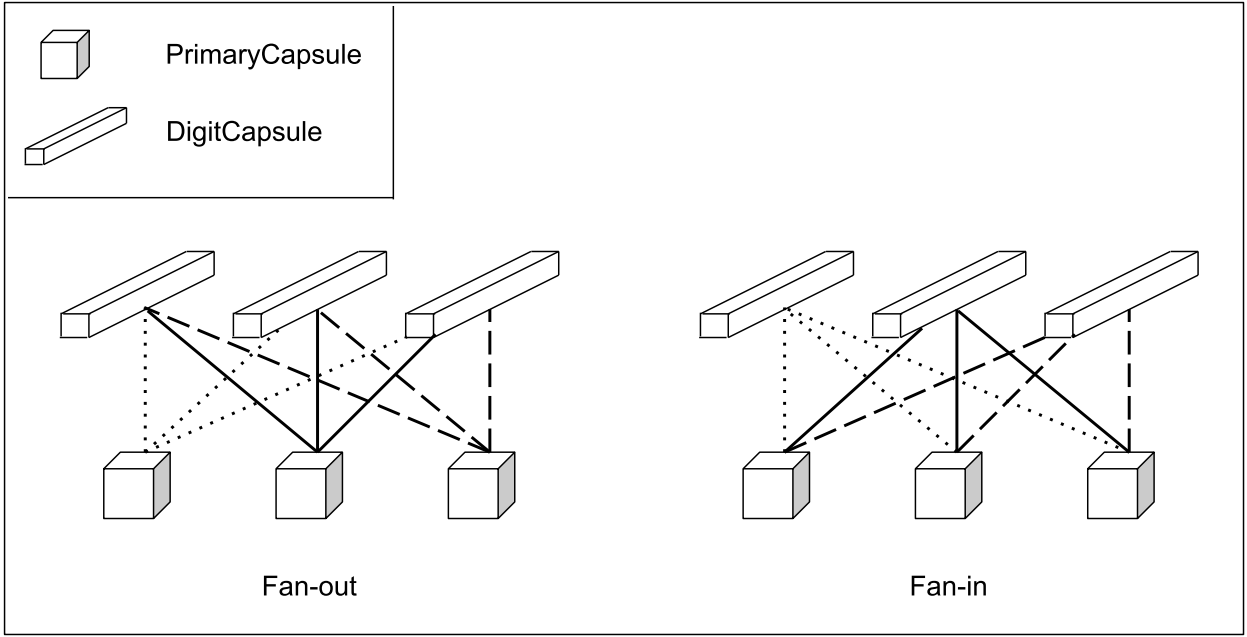}
    \caption{A simplified diagram showing the main difference in Softmax direction between fan-out and fan-in routing. Connections with similar lines are inputs to the same Softmax. Note that the other operations between PrimaryCapsules and DigitsCaps layers are abridged for clarity.}
    \label{fig:softmax}
\end{figure}

\Cref{fig:softmax} shows a simplified diagram highlighting the difference in Softmax calculation direction between fan-out and fan-in variants. All connections with similar line pattern are inputs to the same Softmax. Note how the Softmax is applied across connections fanning into the same DigitsCaps in the fan-in variant, while it is applied across connections fanning out from a single PrimaryCapsule in the fan-out variant.

The accuracy and reconstruction losses are calculated the same way as \citep{Sabour2017}, using margin loss and sum of squared errors loss, respectively.

In the next section, we present the details of our experimental design and the results we obtained.  

\section{Results}

\subsection{PathCapsNet Architecture}

For all of our experiments, each path had the same architecture \cref{tab:path-arch}. The number of paths in each experiment, however, varied and will be clarified for each set of experimental results. We will use the notation PathCapsNet-[num], where [num] is replaced by the number of paths, so PathCapsNet-5 is PathCapsNet with 5 paths. All PrimaryCapsules were 8D with spatial dimensions 7x7. As each path produces one PrimaryCapsule, the number of PrimaryCapsules is equal to the number of paths. The DigitCaps layer was exactly the same as \citet{Sabour2017}. We used 3 routing iterations in all the experiments and whenever we used fan-in routing, we initialized the transformation matrices of the DigitCaps layer randomly from a standard normal distribution. The Adam optimizer was used in all of the experiments using the default parameters and learning rate. When DropCircuit was used, the probability of path dropping was 0.5. Our benchmark was the original CapsNet \citep{Sabour2017} with and without reconstruction and using 3 routing iterations. The benchmark was implemented using the same architecture as reported in the original paper without any modifications, unless otherwise specified. All reported results are based on an average of three trials, with 300 epochs of training each. This number of epochs is relatively small compared to the number used by \citet{Sabour2017}, which seems to be more than 1000 epochs.

\begin{table}[h]
    \centering
    \begin{tabular}{c|c|c|c|c|c}
        Layer & Type & Kernel & Padding & Stride & Output Channels\\
        \hline
        1 & Conv & 9 & 4 & 1 & 16\\
        \hline
        2 & Conv & 9 & 4 & 1 & 16\\
        \hline
        3 & Maxpool & 2 & 0 & 2 & 16\\
        \hline
        4 & Conv & 9 & 4 & 1 & 16\\
        \hline
        5 & Conv & 9 & 4 & 1 & 8\\
        \hline
        6 & Maxpool & 2 & 0 & 2 & 8\\
        \hline
    \end{tabular}
    \caption{Single path architecture}
    \label{tab:path-arch}
\end{table}

\subsection{Experiments on MNIST}

Our experiments were conducted on the MNIST dataset. We trained on 90\% of the training dataset and left 10\% for calculating validation performance, used for selecting the best model. The test performance is reported on the full test set. Following \citet{Sabour2017}, the only augmentation used during training was padding by 2 and random cropping using a 28x28 patch. Our performance results on MNIST are summarized in \cref{tab:mnist-results}. 

\begin{figure}[h]
    \centering
    \includegraphics[width=0.5\textwidth]{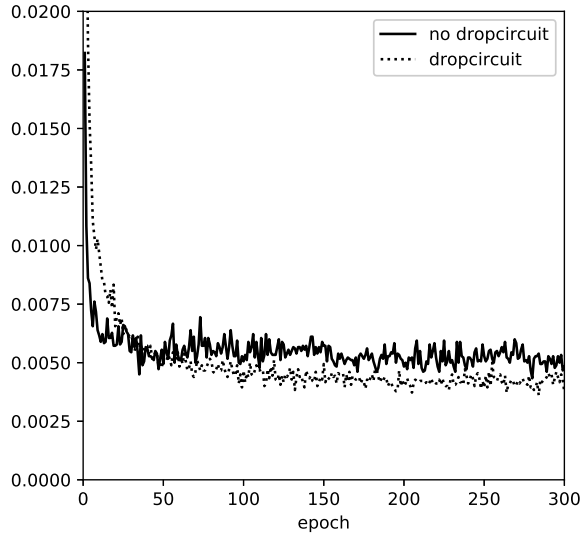}
    \caption{Average validation curves for PathCapsNet-10 with fan-in routing}
    \label{fig:drop-regulz}
\end{figure}

For the no-reconstruction setting, our fan-in routing improved CapsNet test error from 0.48\% to 0.42\%. A similar improvement was observed for PathCapsNet-5, where test error improved from 0.54\% to 0.47\%. With DropCircuit, we observed no improvement for a small number of paths, i.e PathCapsNet-5, while a significant improvement was observed for PathCapsNet-10, where the error improved from 0.52\% to 0.42\%, which is better than the standard CapsNet with only 21\% of the parameters. A regularization effect can be observed from the validation curves \cref{fig:drop-regulz}. For the reconstruction setting, CapsNet had the best validation error of 0.35\%, however, we could achieve a very near performance of 0.38\% with PathCapsNet-16 and DropCircuit with only 44\% of the parameters.

\begin{table}[h]
    \centering
    \begin{tabular}{c|c|c|c|c|c|c|c}
         No. & Architecture & Routing & Paths & DropCircuit & Parameters count & Parameters (\%) & Test error (\%)\\
         \hline
         \multicolumn{8}{l}{\bf{No Reconstruction}}\\
         \hline
         1 & CapsNet & Fan-out & N/A & N/A & 6.8M & 100\% & \SI{0.48 \pm 0.02}{}\\
         \hline
         2 & CapsNet & Fan-in & N/A & N/A & 6.8M & 100\% & \textbf{0.42}\SI{\pm 0.03}{}\\
         \hline
         3 & PathCapsNet & Fan-out & 5 & Yes & 683K & 10\% & \SI{0.54 \pm 0.05}{}\\
         \hline
         4 & PathCapsNet & Fan-in & 5 & No & 683K & 10\% & \SI{0.48 \pm 0.07}{}\\
         \hline
         5 & PathCapsNet & Fan-in & 5 & Yes & 683K & 10\% & \SI{0.47 \pm 0.04}{}\\
         \hline
         6 & PathCapsNet & Fan-in & 10 & No & 1.4M & 21\% & \SI{0.52 \pm 0.03}{}\\
         \hline
         7 & PathCapsNet & Fan-in & 10 & Yes & 1.4M & 21\% & \textbf{0.42}\SI{\pm 0.05}{}\\
         \hline
         \multicolumn{8}{l}{\bf{Reconstruction}}\\
         \hline
         1 & CapsNet & Fan-out & N/A & N/A & 8.2M & 100\% & \textbf{0.35}\SI{\pm 0.04}{}\\
         \hline
         2 & CapsNet & Fan-in & N/A & N/A & 8.2M & 100\% & \SI{0.47 \pm 0.03}{}\\
         \hline
         3 & PathCapsNet & Fan-out & 10 & No & 2.8M & 34\% & \SI{0.44 \pm 0.06}{}\\
         \hline
         4 & PathCapsNet & Fan-in & 10 & No & 2.8M & 34\% & \SI{0.47 \pm 0.02}{}\\
         \hline
         5 & PathCapsNet & Fan-out & 10 & Yes & 2.8M & 34\% & \SI{0.49 \pm 0.02}{}\\
         \hline
         6 & PathCapsNet & Fan-in & 10 & Yes & 2.8M & 34\% & \SI{0.42 \pm 0.05}{}\\
         \hline
         7 & PathCapsNet & Fan-in & 16 & Yes & 3.6M & 44\% & \SI{0.38 \pm 0.02}{}\\
         \hline
    \end{tabular}
    \caption{MNIST results}
    \label{tab:mnist-results}
\end{table}

In the next section we discuss our interpretations and hypotheses explaining the different techniques that contributed to these results and why we think that they enhance the current methodologies.

\section{Discussion}

Three main components, we believe, contributed to the performance of PathCapsNet, namely deep PrimaryCapsules, fan-in routing and DropCircuit. Deep paths, even without fan-in routing and no DropCircuit, could achieve a decent test performance of 0.44\% with only 34\% parameters of the corresponding CapsNet (reconstruction condition 3 in \cref{tab:mnist-results}). We attribute this to the increased representational power of each PrimaryCapsule.

The main rationale behind fan-out routing was that each detected part of an object should contribute more strongly to a single object category rather than to multiple object categories. Fan-in routing, on the other hand, is expressing the idea that for a given object, different detected parts should contribute differently. While both philosophies can be seen to have different pros and cons, making each one optimal for a different set of contexts, empirically, we observed enhancement of generalization when using fan-in routing with DropCircuit on MNIST. One explanatory hypothesis may be that MNIST digits share most of the parts, which means that the problem is not about assigning a part to different digits, but how strongly the different parts contribute to a target digit.

DropCircuit, coupled with fan-in, showed remarkable performance enhancement for conditions with large numbers of paths, and no significant effect for small numbers of paths. We believe DropCircuit, being a form of regularization, needs a sufficiently large number of paths to show a positive effect. With a small number of paths, dropping becomes too destructive, specially with a high dropping rate, to show any significant improvement. We believe DropCircuit, like other drop techniques, is introducing independence between paths and promoting the extraction of more useful PrimaryCapsule representations. 

During experimentation, we noticed that fan-in routing is usually more robust to using DropCircuit, resulting in more enhancement, specially in the reconstruction setting where using DropCircuit worsened the performance of fan-out routing. We think that this is essentially due to the difference in the softmax direction between fan-out and fan-in routing. DropCircuit effectively means that on average only a fraction of PrimaryCaps capsules, and hence prediction vectors contributing to a given DigitCaps capsule, exist at any given iteration. This introduces stochasticity in the fan-out prediction vector sum since the softmax is across DigitCaps capsules. This in effect will make the output of the DigtCaps layer noisy. This noise may not have a great effect on the norm of the capsules, which determines the object identity, but it makes it difficult for the reconstruction layer to function properly. On the other hand, since fan-in routing softmax is local to each DigitCaps capsule, this means that even with missing prediction vectors, on average it can converge to a more stable prediction vector sum and, hence, output.

We could achieve a performance comparable to CapsNet with significant parameter savings. This was possible thanks to a careful coordination of depth, multi-path structure and regularization. Essentially, we substituted the wide convolutional layer of CapsNet with deeper narrower paths regularized by DropCircuit. Fan-in routing enabled the effective utilization of DropCircuit regularization, since our experiments show that fan-out routing is less tolerant to DropCircuit (reconstruction cases 3 and 5 in \cref{tab:mnist-results}). Another component that contributed to reducing parameter counts was max-pooling which may be considered incompatible with CapsNet and its equivariance aim. However, we showed that it is possible to use max-pooling layers in PathCapsNet, which allowed further parameter savings without sacrificing performance. 

\begin{figure}
    \centering
    \includegraphics[width=\textwidth]{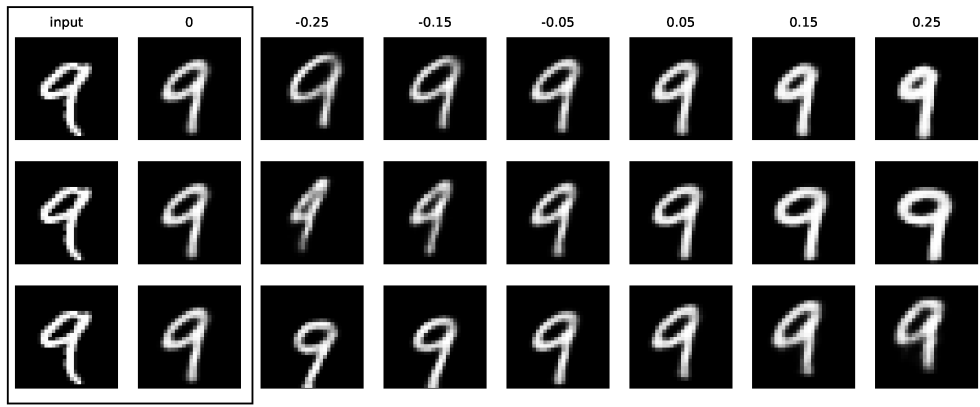}
    \caption{Perturbing different dimensions of PathCapsNet-10 (DropCircuit and fan-in). The images in the box are, from the left, the input and the unperturbed reconstruction, respectively.}
    \label{fig:pcnet-reconst}
\end{figure}

Moreover, experiments perturbing different dimensions in the DigitCaps layer confirmed that, even when using max-pooling, different pose parameters can be successfully learned \cref{fig:pcnet-reconst}. For example, perturbing the first three dimensions of the DigitCaps layer of one of the models \cref{fig:pcnet-reconst} suggested that the first dimension in the model was controlling multiple pose parameters, like the elongation of the circular and linear regions and stroke thickness. The second dimension seemed to affect the circle axes orientations and also stroke thickness, while the third dimension resulted in a combination of vertical translation, vertical axis inclination and elongation of the circular part.

\section{Conclusion}

We have introduced PathCapsNet, a multipath capsule network that can achieve better or comparable performance to CapsNet with significant parameter savings. In order to achieve this, we used regularization by DropCircuit along with a new variant of dynamic routing by agreement, fan-in routing. The careful coordination of depth, max-pooling, fan-in routing and DropCircuit allowed for maintaining CapsNet performance, while cutting down parameter counts considerably. Reconstructions with perturbations showed that the use of max-pooling is not necessarily in conflict with retaining location information and pose estimation. The independence of paths renders the model suitable for model parallelism, a property which we didn't investigate in detail and we leave for future work. We think there is still more space for enhancing PathCapsNet, specially in the reconstruction setting where we believe there is a complex interaction between routing, DropCircuit and reconstruction.

\bibliographystyle{plainnat}
\bibliography{bib}

\end{document}